\documentclass[11pt]{article}
\usepackage[final]{acl}
\usepackage{times}
\usepackage{latexsym}
\usepackage[T1]{fontenc}
\usepackage[utf8]{inputenc}
\usepackage{microtype}
\usepackage{booktabs}
\usepackage{graphicx}
\usepackage{amsmath}
\usepackage{amssymb}
\usepackage{tabularx}
\usepackage{array}
\usepackage{tikz}
\usetikzlibrary{arrows.meta,positioning}
\title{From Monolithic Blending to Agentic Orchestration: Dynamic Response for
Conversational Assistants at Scale}
\author{
  \textbf{Cen Mia Zhao} \quad \textbf{Peng Wang} \quad \textbf{Chuan Shi} \quad \textbf{Yufeng Zhang} \quad \textbf{Ying Lyu} \\
  \textbf{Wanmeng Ren} \quad \textbf{Robert Xue} \quad \textbf{Claire Na Cheng} \quad \textbf{Yashar Mehdad} \\
  Airbnb, Inc. \\
  \texttt{\{mia.zhao, peng.wang, chuan.shi, wayne.zhang, ying.lyu,} \\
  \texttt{emma.ren, robert.xue, claire.cheng, yashar.mehdad\}@airbnb.com}
}
\begin{document}
\maketitle
\begin{abstract}
Conversational assistants can blend retrieval, action selection, escalation,
and wording in one model path, or separate those roles. We report a production migration of a conversational assistant for customer support at a large
accommodation marketplace (millions of conversations per month, 11 languages,
10-second P90). Dynamic Response (DR) replaces a single Qwen3-235B-A22B blended
responder with a bounded ReAct orchestrator over typed tools plus a smaller
generator that writes from a backend-validated context contract.
Because the
migration also changed prompts, alignment, and serving, we attribute each effect
to its cause and claim as architecture effects only those measured on identical
replayed turns: typed entity selection moves the reservation selector to a
precision-first operating point (precision 8.3\% to 89.1\% at recall 75.2\% to
67.3\%), and typed action IDs with a membership check remove observed
structured-action hallucination (2.14\% to 0.0\%). A low-ramp A/B test
reproduces the replay escalation reductions: hard-escalation responses fall
5.60\% to 3.08\% and soft-escalation responses 9.56\% to 2.49\% while
production handoff volume holds roughly steady, and self-solve is directional
(+5.1 points, 95\% CI $[-2,+12]$). Serving optimizations cut orchestrator P90
from 3.87s to 2.24s on a GPU footprint reduced by roughly one-third, and
self-hosting reduces estimated annual model-serving cost by more than an order
of magnitude (Appendix~\ref{app:cost}).
\end{abstract}
\section{Introduction}
Conversational assistants built on LLMs increasingly act as first-line
customer-support agents. In our setting, the
assistant answers policy questions, selects structured cancellation and refund
flows, asks clarifying questions, and routes to humans when automation should
stop. Each turn must fit a 10-second P90 end-to-end budget, including translation
across 11 languages.
The central systems problem is that support turns require both \emph{decision
making} and \emph{wording}. A single blended responder can retrieve context,
select entities or actions, decide whether to clarify or escalate, and write the
reply, but these responsibilities fail differently. Wrong entity cards,
unsupported action language, or premature human handoff become hard to localize
when routing and prose are produced in one generation path.
Dynamic Response (DR) separates those roles. A large orchestrator plans, calls
typed tools, and owns routing, action, and escalation decisions; a smaller
generator writes the user-facing reply from a backend-validated context contract.
We focus on systems design; scoring-instrument construction, offline-to-online
calibration, and launch-decision methodology are companion evaluation work.
Because the migration changed architecture, prompts, alignment, and serving
together, we attribute each reported effect to its cause and claim as
architecture effects only those measured on identical replayed turns
(Table~\ref{tab:attribution}); the remaining results are system-migration
outcomes.
We make four contributions. Unlike a RAG workflow with an API wrapper, DR
structures and logs intermediate decisions, constrains generation to orchestrator
decisions and retrieved evidence, and uses role-specific model sizes. The
transferable principle is to turn high-consequence decisions into explicit,
backend-validated typed tools and keep wording downstream and constrained.
Concretely: (1)~\textbf{production decomposition}, a typed context contract that
separates orchestration from final prose; (2)~a \textbf{tool-mediated runtime}
with bounded ReAct planning, typed schemas, boundary guardrails, and conservative
fallback; (3)~a \textbf{serving recipe}, a self-hosted open-weight MoE on vLLM
with tensor-parallel serving, FlashInfer, EAGLE-3 speculative decoding, and a
latency-fit layout; and (4)~\textbf{operational lessons} on context handoff,
tool false positives, soft escalation, and open-weight cost and control trade-offs.
\section{Background and Problem Setting}
\subsection{Task}
When a guest or host starts support, the assistant must understand the request,
retrieve user state (reservation, listing, payment, account), retrieve policy
and help-center evidence, and choose among structured action cards,
clarification, informational answers, and human routing. Many turns require
cross-source reasoning: a cancellation may depend on reservation dates, listing
policy, payment state, and regional rules. The system must also support
translation, avoid unsupported actions, keep references traceable, and preserve
human handoff; these constraints make architecture as important as base-model
choice.
\subsection{Monolithic Blending Baseline}
The predecessor used the same Qwen3-235B-A22B-class model as a monolithic
responder. Backend services preassembled candidate help or policy snippets,
user-state variables, and renderable structured-action candidates. A single
blended model path then chose the response strategy, decided whether to present an
action or escalate, and wrote the user-facing reply; backend rules validated
renderability, blocked unsupported outputs, and provided conservative fallback.
This design was simple and avoided a second model call, but it entangled
\emph{what to do} with \emph{what to say}: wrong actions, unsupported handoff
language, or wrong entity references all appeared in one generation trace, hard to
assign to retrieval, routing, or wording. Adding tools also enlarged the blended
prompt, since every capability had to be exposed to the path that produced final
prose. DR keeps the same large-model reasoning capacity for planning while making
routing, tool use, and final wording separately observable and controllable. A
stronger monolithic system with constrained decoding or structured JSON output
would enforce syntactic validity, but routing and prose would still share one
generation trace that cannot be individually localized, gated, or rolled back,
and role-specific model sizing would remain unavailable.
\subsection{Related Work}
DR combines ideas from RAG, tool-using agents, and LLM serving. RAG
\citep{lewis2020rag} grounds generation in retrieved evidence, but single-pass
RAG-style support systems can still couple evidence selection, action choice,
escalation, and final wording. Building on chain-of-thought prompting
\citep{wei2022cot}, ReAct \citep{yao2023react} interleaves reasoning and acting,
which maps naturally to support resolution as tool calls plus observations, and
tool-augmented LLMs \citep{schick2023toolformer,qin2024toolllm} study how models
select and use external tools.
Serving efficiency also matters because orchestration must fit a production
latency budget: mixture-of-experts models \citep{shazeer2017moe}, grouped-query
attention \citep{ainslie2023gqa}, vLLM \citep{kwon2023vllm}, and speculative
decoding \citep{leviathan2023speculative,chen2023speculative} make large routing
models practical to serve. Agent frameworks such as AutoGen \citep{autogen2023},
LangGraph \citep{langgraph}, Smolagents \citep{smolagents}, and CrewAI
\citep{crewai} provide useful orchestration abstractions, but our deployment
required backend-enforced control over routing, guardrail placement, logging,
context handoff, and fallback behavior.
Closest to our setting are recent LLM customer-support agents: benchmarks for
resolving real e-commerce issues \citep{zhang2025ecombench} and for asking
clarifying questions in conversational product search \citep{chen2025productagent},
and deployed systems built around continuous-improvement data flywheels
\citep{mehdad2025flywheel} and incremental summarization with agent feedback
\citep{mehdad2025summarization}. These target task performance, data loops, and
assistance features, and typically evaluate a single end-to-end policy. DR
instead turns high-consequence decisions into typed, backend-validated tool
calls and reports the migration with explicit effect attribution on identical
replayed turns; the evaluation methodology behind our release gates is
companion work.
\section{Dynamic Response Architecture}
\label{sec:dr}
\subsection{Design Principles and Runtime}
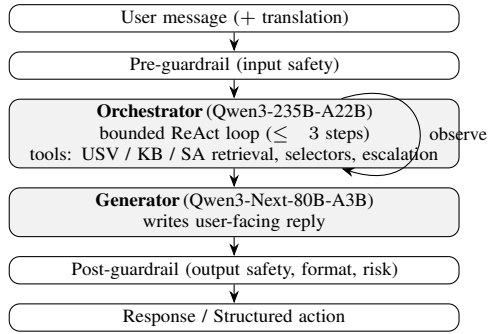
\begin{figure}[t]
\centering
\begin{tikzpicture}[
  node distance=2.5mm,
  box/.style={draw, rounded corners, align=center, inner sep=2pt,
              text width=5.8cm, font=\scriptsize},
  >={Stealth[]}
]
\node[box] (in) {User message ($+$ translation)};
\node[box, below=of in] (pre) {Pre-guardrail (input safety)};
\node[box, below=of pre, fill=gray!10] (orch)
  {\textbf{Orchestrator}\,(Qwen3-235B-A22B)\\ bounded ReAct loop ($\leq 3$ steps)\\
   tools: USV / KB / SA retrieval, selectors, escalation};
\node[box, below=of orch, fill=gray!10] (gen)
  {\textbf{Generator}\,(Qwen3-Next-80B-A3B)\\ writes user-facing reply};
\node[box, below=of gen] (post) {Post-guardrail (output safety, format, risk)};
\node[box, below=of post] (out) {Response / Structured action};
\draw[->] (in) -- (pre);
\draw[->] (pre) -- (orch);
\draw[->] (orch) -- (gen);
\draw[->] (gen) -- (post);
\draw[->] (post) -- (out);
\draw[->] (orch) to[out=18,in=-18,looseness=4]
  node[right,font=\scriptsize]{observe} (orch);
\end{tikzpicture}
\caption{DR runtime. The large orchestrator plans and calls tools in a bounded
ReAct loop; the smaller generator writes from the typed contract. Guardrails wrap
input, tool use, and output.}
\label{fig:arch}
\end{figure}

DR follows three design rules. \textbf{Separate routing from wording:} the
orchestrator chooses tools, actions, and escalation; the generator writes the
reply without overriding those decisions. \textbf{Retrieve on demand:} the
orchestrator requests only the context the turn needs and passes a curated state
to the generator. \textbf{Make control flow observable:} each orchestration
step is logged as a structured event, so failures can be localized and
guardrails attached at the right boundary.
Figure~\ref{fig:arch} shows the runtime. After input safety checks and
translation, the orchestrator enters a bounded ReAct loop capped at three steps.
It may call retrieval tools, selector tools, the escalation tool, or response
generation; tool observations are appended to orchestration state. Once response
generation is selected, the generator receives the typed context contract and
writes the final reply. Post-generation checks validate formatting, references,
unsupported actions, and safety-sensitive output before delivery.
\subsection{Model Roles}
The orchestrator is a fine-tuned open-weight MoE model, Qwen3-235B-A22B
\citep{qwen2025qwen3}, with 235B total and 22B active parameters. It is the same
large-model class used by the monolithic baseline, but DR restricts it to
structured tool-call decisions, terminal actions, and fallback conditions rather
than user-facing prose. The generator is Qwen3-Next-80B-A3B, selected because
final responses are longer and must fit the latency budget. It writes from the
contract in Section~\ref{sec:contract} and cannot select a new action, call
retrieval, or initiate escalation.
\subsection{Tool Registry}
The orchestrator uses a typed registry (full schema in
Appendix~\ref{app:components}, Table~\ref{tab:tools}). Each entry defines schema,
eligibility, logging fields, and guardrail hooks. User-facing tools change the
customer experience; internal tools retrieve context or constrain the action
space. Tool outputs are small typed objects when possible, letting the backend
validate IDs, reason codes, and source membership before generation.
\subsection{Two-Phase Deployment}
DR shipped in two stages, both retaining fallback to the previous path to
separate migration risks. Stage 1 used one orchestrator call over pre-fetched
context, preserving much of the backend while validating structured routing;
unlike Monolithic, it invoked a separate generator instead of writing final prose.
Stage 2 introduced the full tool-mediated loop, letting the orchestrator decide
which retrieval tools to call, whether entity selection is needed, whether a
structured action applies, and when generation should run. The
architecture-local wins reported below come from Stage 2's typed decisions;
Stage 1 mainly validated structured routing and the handoff, and a per-stage
ablation is out of scope.
\section{Context Contract, Guardrails, and Fallback}
\label{sec:contract}
\subsection{Context Contract}
The handoff between orchestrator and generator is a product interface, not a
prompt detail. The forward contract passes task state, eligible tools, retrieved
sources, user-state variables, policy constraints, locale, and prior
observations. The return contract contains response text, source references, an
optional orchestrator-selected action-card reference, and rendering fields; full
fields are in Appendix~\ref{app:ctx}.
Two constraints are central. First, the generator may reference only tools,
sources, and user-state keys present in the contract, and post-generation checks
enforce this closure. Second, the handoff includes the orchestration state needed
for coherence, not only cited snippets: cited-only handoff was shorter but lost
context such as why an action was unavailable.
\subsection{Guardrails and Fallback}
DR places guardrails before orchestration, inside the ReAct loop, and after
generation. Pre-checks classify input safety and eligibility; mid-stream checks
validate tool schema, eligibility, and consistency with prior state; post-checks
validate references, action-card IDs, format, and risk-sensitive phrasing. If the
orchestrator exceeds the iteration limit, emits an invalid tool call, or receives
inconsistent observations, DR falls back to conservative generation over available
context. Fallback cannot invent actions or bypass safety routing; every
fallback logs its trigger, separating infrastructure failures (timeout,
malformed tool results) from policy failures (no eligible action, ambiguous
state), which require different owners.
\section{Model and Serving Choices}
\label{sec:serving}
\subsection{Open-Weight Serving}
The orchestrator and generator are self-hosted on vLLM \citep{kwon2023vllm}.
Open weights let us align models to policy with full-parameter preference
optimization (Appendix~\ref{app:alignment}), keep pre-production and production
settings identical, tune serving to our traffic shape, and run large shadow
tests; reward calibration to production self-solve is out of scope. At
our volume and decoding settings, self-hosting reduced estimated annual
model-serving cost by more than an order of magnitude relative to a
proprietary-API baseline; these are internal order-of-magnitude estimates
whose basis, scope, and exclusions are detailed in Appendix~\ref{app:cost}.
\subsection{Inference-Engine Tuning}
The orchestrator runs deterministic decoding on H100 GPUs, where latency rather
than throughput was the binding constraint. Switching from tensor-plus-expert
parallelism to tensor-parallel-only serving \citep{shoeybi2019megatron}, enabling
asynchronous scheduling, and using FlashInfer kernels \citep{ye2025flashinfer}
reduced P90 from 3.87s to 2.24s while shrinking the GPU footprint by roughly
one-third. Orchestrator-only serving latency across operating points is reported in
Appendix~\ref{app:model} (Table~\ref{tab:latency}), and the end-to-end
latency distribution by stage is reported in Table~\ref{tab:e2e-latency}. EAGLE-3
speculative decoding \citep{li2025eagle3} is being validated as a serving-only
optimization toward a further-reduced operating point, and orchestrator and
generator capacity are co-located behind the stable contract; details are in
Appendix~\ref{app:model}.
\begin{table}[t]
\centering\small
\setlength{\tabcolsep}{6pt}
\begin{tabular}{@{}lccc@{}}
\toprule
Stage & P50 & P90 & P99 \\
\midrule
Translation & 0.043 & 0.059 & 0.792 \\
Orchestration (incl.\ tools) & 2.772 & 4.151 & 7.404 \\
Generation & 0.926 & 1.325 & 1.695 \\
Guardrails + rendering & 0.734 & 1.460 & 2.292 \\
\midrule
End-to-end & 4.787 & 6.539 & 10.386 \\
\bottomrule
\end{tabular}
\caption{End-to-end latency distribution by stage (seconds), DR arm on live
traffic over the stabilized tail 2026-03-02 to 2026-03-05 ($n = 4{,}155$ turns).
Percentiles are computed over per-turn stage sums, not as sums of per-stage
percentiles, so the stage rows need not add to the end-to-end row; the
unattributed residual is at most 0.39s at P99. The orchestration stage spans the
full bounded ReAct loop, including tool executions and up to three model calls,
so it exceeds the single-call orchestrator serving latency in
Table~\ref{tab:latency}. End-to-end P90 (6.539s) is within the 10s release gate,
and orchestration is the dominant stage (about 58\% of the median turn).}
\label{tab:e2e-latency}
\end{table}
\section{Controlled Comparison and Launch Process}
\label{sec:comparison}
We evaluate DR as an architecture migration, with evidence from zero-exposure
shadow replay, a low-ramp online A/B experiment, and targeted human and risk
review, all under fixed production release checks.
\paragraph{What we attribute to architecture.}
The migration changed the architecture, generator placement, prompt and tool
schemas, alignment, and serving stack. The high-level decision-model class
remained Qwen3-235B-A22B-class in both systems: Monolithic used it as a blended
final responder, while DR uses it as the orchestrator. We therefore separate
\emph{architecture-local} effects, which correspond to typed selector, action,
and escalation decisions, from \emph{model and alignment} and serving effects,
which would also benefit a single-call system. Table~\ref{tab:attribution}
states this mapping. We do not claim a pure
architecture-only ablation; the end-to-end comparison is a production
migration, with component replay localizing the main behavior changes.
\begin{table*}[t]
\centering\small
\setlength{\tabcolsep}{6pt}
\begin{tabular}{@{}lll@{}}
\toprule
Effect & Source & Attributable to \\
\midrule
Selector precision 8.3 to 89.1\% & explicit selector tool & architecture \\
SA hallucination 2.14 to 0\% & typed action IDs + check & architecture \\
Soft-escalation responses $-74$\% & escalation as a tool & architecture \\
Core correctness (alignment effect) 0.825 to 0.93 & self-host + alignment & model and alignment \\
P90 $-42$\%, GPU footprint $-33\%$ & inference-engine tuning & serving \\
\bottomrule
\end{tabular}
\caption{Attribution of the main gains. Architecture-local effects are measured
on identical replayed turns; model, alignment, and serving effects would also
benefit a single-call system. \emph{Core correctness (alignment effect)} is the
self-hosting-plus-alignment gain on one instrument (0.825 pre-alignment baseline
to 0.93 aligned model); it is a different measurement from \emph{shadow-window
logic correctness} in Table~\ref{tab:turn}, where both arms already use the
aligned self-hosted models.}
\label{tab:attribution}
\end{table*}
\paragraph{Evaluation scope.}
Table~\ref{tab:evidence_scope} (Appendix~\ref{app:err}) maps evidence to claims:
replay localizes component failures on identical turns, human and risk review
gates broadening, and A/B measures real-user outcomes under identical
eligibility. We claim only architecture-local behavior, release-gated response
quality, low-ramp online outcomes, and serving cost and latency; evaluator
construction, judge certification, and offline-to-online calibration are
companion work. Fixed launch
gates were: bad-response rate $\leq 1\%$, correctness $\geq 90\%$, groundedness
$\geq 75\%$, structured-action FPP $\leq 7\%$, structured-action recall $\geq
60\%$, and end-to-end P90 $\leq 10$s. Correctness and groundedness in
Table~\ref{tab:turn} come from fixed production release-check judges
\citep{zheng2023judge} on a calibrated $\approx$1k-turn set. Each judge is certified against expert-adjudicated human
labels before use (the problem-solution judge reaches ICC 0.88 and AUC 0.84
against a 0.75 human baseline). These scores interpret the architecture
comparison and are not treated as powered significance tests.
\subsection{Shadow Replay}
\label{sec:shadow_replay}
We replayed production requests through both systems while serving only
Monolithic responses. Fixed user behavior plus logged inputs, tool decisions, and
outputs lets us localize failures to orchestration, tool selection, or final
wording.
\paragraph{Response quality and risk.}
Table~\ref{tab:turn} shows that DR remains within production release thresholds
while shifting the error profile. Orchestrator-decided metrics (selector,
escalation, and reference checks) and generator-realized metrics (correctness,
groundedness, style) are reported in separate columns, so each win or regression
is attributed to the role that produced it. The main architecture wins are zero observed structured-action hallucinations
and lower user-role confusion; the main regressions are generator-side USV
attribution, style and tone, and repetition. USV errors mix strict citation mismatch with
genuine gap filling, so we are moving from per-sentence to per-key attribution to
separate those cases (Appendix~\ref{app:err}). Style and repetition regressions
are treated as generator headroom rather than evidence against the split, because
upstream decision quality improves
(Appendices~\ref{app:generator-prompt-headroom}
and~\ref{app:generator-model-selection}). The structured-action hallucination
claim is scoped to structured-action references only, and the 0.0\% is the
delivered rate: the generator can reference only actions present in the
contract, the backend independently blocks any spurious action ID before
delivery, and blocked attempts are logged separately.
\begin{table}[t]
\centering\small
\setlength{\tabcolsep}{4pt}
\renewcommand{\arraystretch}{0.92}
\begin{tabular}{@{}p{0.42\columnwidth}ccc@{}}
\toprule
Metric & DR Orch. & DR Gen. & Monolithic \\
\midrule
Shadow-window logic correctness ($\uparrow$)\textsuperscript{$\dagger$} & n/a & 91.69 & \textbf{92.62} \\
Groundedness ($\uparrow$) & n/a & 87.62 & \textbf{89.82} \\
Problem solution ($\uparrow$) & n/a & \textbf{84.36} & 83.91 \\
Style and tone ($\uparrow$) & n/a & 68.50 & \textbf{76.20} \\
User-role confusion ($\downarrow$) & n/a & \textbf{1.91} & 2.76 \\
Repeated pattern ($\downarrow$) & n/a & 32.54 & \textbf{27.73} \\
SA hallucination ($\downarrow$) & \textbf{0.00} & \textbf{0.00} & 2.14 \\
KB hallucination ($\downarrow$) & 0.06 & \textbf{0.00} & 0.03 \\
USV attribution error ($\downarrow$) & \textbf{0.00} & 6.19 & 0.12 \\
Selector correctness ($\uparrow$) & \textbf{93.43} & n/a & 91.32 \\
Escalation correctness ($\uparrow$) & \textbf{97.30} & n/a & 96.30 \\
\bottomrule
\end{tabular}
\caption{Turn-level quality on a fixed shadow window ($n \approx 1{,}000$
turns). \textbf{DR Orch.} reports metrics decided by the orchestrator (reference
checks, selector, escalation); \textbf{DR Gen.} reports metrics realized in the
generator's final wording; ``n/a'' marks a non-applicable role. Arrows give the
preferred direction; bold marks the better arm.
\textsuperscript{$\dagger$}\emph{Shadow-window logic correctness} compares final
wording on identical replayed turns and is a different instrument from
\emph{core correctness (alignment effect)} in Table~\ref{tab:attribution}.}
\label{tab:turn}
\end{table}
\paragraph{Entity selection.}
The selector shifts to a precision-first operating point (full precision and
recall in Table~\ref{tab:selector}). Monolithic favored
recall and surfaced many wrong entity cards; DR makes entity choice an explicit
typed orchestrator decision. The design goal of the precision-first point is
that a selector miss should not surface a wrong entity card, which would be
misinformation. On the shadow slice where the Monolithic arm surfaced an entity
but the DR selector returned null ($n = 6{,}945$ turns, drawn from the full
replay traffic), that goal largely holds. Here \emph{wrong entity} counts a
delivered response that references an incorrect specific entity --- a stricter
event than the spurious-card false positives that dominate Monolithic's
precision in Table~\ref{tab:selector}: DR references a wrong entity in only
1.01\% of these turns (Monolithic 0.20\% on the same turns), so a miss rarely
produces entity-level misinformation. The cost is
elsewhere in resolution quality. On this slice DR gives an entity-free answer
judged correct in 47.75\% of turns, asks a clarifying question in 1.43\%, and
escalates in 3.87\%; the remaining turns give entity-free answers not judged
fully correct. This slice is a known weakness: picker correctness here (about
48\%) is well below the selector's overall operating point, and raising
miss-slice resolution is ongoing work. Precision is also enforced structurally:
the backend rejects any rendered card whose ID the orchestrator did not select.
\begin{table}[t]
\centering\small
\setlength{\tabcolsep}{4pt}
\begin{tabular}{lcccc}
\toprule
 & \multicolumn{2}{c}{Reservation} & \multicolumn{2}{c}{Listing} \\
\cmidrule(lr){2-3}\cmidrule(lr){4-5}
System & Prec. & Rec. & Prec. & Rec. \\
\midrule
Monolithic & 8.3 & 75.2 & 9.0 & 87.7 \\
DR (no selector ctx) & 82.1 & 45.6 & 66.8 & 48.9 \\
DR (+ ctx, v2) & 87.5 & 58.5 & 71.7 & 55.2 \\
DR (+ ctx, v3) & \textbf{89.1} & 67.3 & \textbf{73.3} & 58.5 \\
\bottomrule
\end{tabular}
\caption{Selector precision and recall (\%, approximately 1k shadow samples).}
\label{tab:selector}
\end{table}
\paragraph{Structured-action selection.}
DR exposes the precision and recall trade-off in the orchestrator prompt
(Appendix~\ref{app:err}, Table~\ref{tab:ac}). We distinguish the F1 frontier
from the shipped point: one configuration reaches Monolithic-level F1 but exceeds
the false-positive gate, while the deployed configuration sacrifices a small
amount of F1 to satisfy the false-positive and recall gates. Thus the shipped
claim is gate compliance; the parity claim refers only to the achievable frontier.
\paragraph{Escalation behavior.}
Treating escalation as a tool gives a clearer control point than allowing handoff
phrases to appear in final prose. DR reduces avoidable soft escalation, while
replay shows that nearly all Monolithic hard escalations remain hard-escalated,
so the reduction is not simply suppressed routing. Per-risk-category risk
prevalence on the same replayed traffic is reported in
Table~\ref{tab:risk-underesc}, so the escalation
reduction can be checked against risk exposure rather than taken on faith.
\subsection{Human Review and Risk Monitoring}
\label{sec:human_risk}
Pre-broadening review covered 1{,}173 DR conversations and 2{,}335 dual-labeled
judgments from 28 annotators. One development snapshot passed correctness and
groundedness checks but failed an internal risk gate, blocking broadening. We
include this as release-process evidence, not as a launch-ready result. Error
modes appear in Appendix~\ref{app:err}.
We also compared Monolithic and DR on false promise, medical-records, tax,
ID-verification, and refund risks. Review found no statistically reliable
increase, though this is monitoring evidence rather than proof of equivalence
(Table~\ref{tab:risk-underesc}). In internal red-team exercises
\citep{perez2022redteam,ganguli2022redteam}, DR introduced no qualitatively new
attack surface: of the issues identified in manual red-teaming of DR, all that
were retested (24 of 24, across 123 attempts by six testers) reproduced at
least once on the Monolithic production system, and in automated adversarial
template scans DR's pass rate matched or exceeded Monolithic's on all three
primary templates after iterative remediation. Findings follow standard
remediation.
\begin{table}[t]
\centering\small
\setlength{\tabcolsep}{4pt}
\begin{tabular}{lcc}
\toprule
 & \multicolumn{2}{c}{Risk prevalence, \% [95\% CI]} \\
\cmidrule(lr){2-3}
Risk category & Monolithic & DR \\
\midrule
Cancellation / refund & 7.6 [6.7, 8.7] & 8.4 [7.3, 9.6] \\
General & 4.0 [3.3, 4.8] & 4.1 [3.3, 4.9] \\
False promise & 1.2 [0.8, 1.7] & 1.2 [0.8, 1.7] \\
Tax & 1.0 [0.7, 1.4] & 0.9 [0.6, 1.4] \\
Medical records & 0.0 [0.0, 0.2] & 0.1 [0.0, 0.3] \\
\bottomrule
\end{tabular}
\caption{Per-risk-category risk prevalence on identical shadow traffic
(Monolithic $n{=}2{,}795$ turns; DR $n{=}2{,}318$ responses: prevalence is
computed among responses because a subset of replayed turns routed to typed
non-response actions such as entity pickers, which would otherwise bias the
comparison downward). Flags come from five risk classifiers calibrated against
subject-matter-expert labels, built as a three-stage hierarchical pipeline
(category filter, sub-skill filter, focused per-pattern scoring with
policy-encoded prompts). No category shows a statistically significant
difference between the systems (all 95\% CIs overlap). ID-verification and
other safety-sensitive contact reasons are routed deterministically to humans,
independent of the model's decision (Section~\ref{sec:human_risk}).}
\label{tab:risk-underesc}
\end{table}
\subsection{Online A/B (Low Ramp)}
\label{sec:ab}
We then ran a low-ramp online A/B test under identical eligibility, randomized
at the user level (sessions appear later only as an analysis unit, in the
session-level results of Table~\ref{tab:per-language}, not as the
randomization unit), and report the 5\% ramp stage at its final day (2026-03-05;
1{,}328 control and 1{,}346 treatment users enrolled). Escalation rates are
turn-level response rates over all turns from enrolled users, with the same
definitions in both arms: a hard-escalation response routes the user to a
human agent after a user-initiated request (excluding guardrail-triggered
routing), and a soft-escalation response offers to connect the user to a
human. The escalation reductions observed in replay reproduce online:
hard escalation falls from 5.60\% to 3.08\% (95\% CI on the change
$[-4.1, -0.9]$ points) and soft escalation from 9.56\% to 2.49\% (95\% CI
$[-8.1, -5.8]$ points). The production human-handoff rate on the same cohort
held roughly steady (8.56\% to 8.20\%), so the reduction reflects fewer
escalation-framed responses, not fewer human handoffs --- itself evidence
against under-escalation. Self-solve moves
directionally (+5.1 points, 95\% CI $[-2,+12]$) but is not significant at this
ramp; contact-rate effects are likewise directional.
At the fuller ramp the session-level self-solve gain is smaller but well
powered; we treat the low-ramp estimate as directional and escalation
behavior, not self-solve, as the supported online result. Per-language and
per-tier splits are in Appendix~\ref{app:err}, Table~\ref{tab:per-language}
(different windows and denominators; the wider window reproduces the
reductions, e.g., English soft escalation 9.43\% to 2.03\%).
\section{Production Rollout and Operations}
\label{sec:rollout}
DR rolled out in stages: zero-exposure replay while the previous system served
responses; low-traffic canary with immediate fallback; expansion by region and
language; and retained rollback to the previous system during migration. Because
the orchestrator, generator, tool calls, and post-generation checks are logged
separately, operators can localize failures to a boundary and roll back only the
affected component.
Operations follow a weekly release train: tool or policy changes update the
registry and prompt contracts first, and model updates stage behind the same
interface, so generator swaps do not change the orchestrator and prompt updates
do not change the backend renderer. Training and monitoring data is
de-identified and excludes opt-out users.
\section{Lessons Learned}
\label{sec:lessons}
\paragraph{The handoff is the architecture.}
Decomposition only works when the boundary between orchestrator and generator is
explicit. Early prototypes treated the handoff as a prompt string, making
failures hard to localize; the typed contract made the boundary inspectable and
gave guardrails concrete objects to validate.
\paragraph{High-impact decisions should be explicit tools.}
Escalation, entity selection, and action-card rendering are too consequential to
leave to final prose: a single-call model can say ``an agent can help'' without
making a routing decision, or surface a wrong entity card when clarification
would be safer. DR makes these orchestrator tool decisions; the generator may
explain the selected path but cannot create a new escalation or action through
wording alone.
\paragraph{Role separation changes the optimization target.}
Because the generator no longer performs global planning, it can be smaller,
provided the contract carries the evidence needed for grounded wording and
prevents action changes. Open weights made the split practical: self-hosting let
us tune parallelism, kernels, speculative decoding, and placement while keeping
the previous system as a large-scale fallback during migration.
\section{Conclusion}
We presented Dynamic Response, a two-model agentic architecture for production
conversational assistants: a large orchestrator plans and calls tools, and a
smaller generator writes the final reply. Self-hosted open-weight MoE models and
serving optimizations fit a 10-second P90 multilingual budget with observable,
modular control flow. The added complexity pays off when the handoff contract,
guardrails, and serving stack are first-class parts of the architecture.
\section*{Limitations}
DR was built for one support domain with mature tools, stable policy sources, and a
human-support fallback. Decomposition is most likely to pay off when
high-consequence typed decisions must be logged, validated, and rolled back,
when the latency budget can absorb a second model call, and when tools and
policy are mature; it helps less when tools are sparse, when policy changes
faster than the contract and registry can be maintained, or when wrong decisions
carry little risk. It also has a real operational price: a second serving stack
and release train, and version coordination across the tool registry,
orchestrator prompts, and backend validators. The second model call adds
operational complexity even though serving optimizations keep latency within
budget. Cost and latency figures reflect our hardware, traffic, and decoding
settings.
\paragraph{Reproducibility.} The paper releases the reusable interfaces (the
context-contract schemas and invariants, the tool-registry schema, and the
release-evidence framework); the shadow sets, judge prompts, and labeled data are
proprietary and cannot be released.
This paper reports
release-process evidence needed to interpret the architecture, but omits the
companion methodology used to construct evaluator roles, certify LLM judges,
calibrate offline scores to online outcomes, or select prompt and model variants.
The online A/B evidence is low-ramp migration evidence; self-solve and contact
effects are directional at the low ramp, while the fuller-ramp session-level
self-solve gain is smaller but powered (Table~\ref{tab:per-language}). Because DR was
evaluated as a production migration, the comparison should not be read as a pure
causal ablation of architecture alone.
\section*{Ethical Considerations}
The system processes customer-support conversations. Data used for training,
debugging, and monitoring is de-identified, opt-out users are excluded before
training use, and log access is restricted; human annotators reviewed only
de-identified conversations. Pre-, mid-, and post-generation
guardrails reduce harmful, discriminatory, privacy-violating, or policy-violating
outputs. Human escalation remains available, and safety-sensitive topics can be
routed automatically. We also monitor under-escalation because reducing escalations
is only desirable when automation remains appropriate; escalation is a typed tool with reason codes, so under-escalation
is monitorable per risk category and per language in production.
Table~\ref{tab:risk-underesc} reports per-category risk prevalence,
Table~\ref{tab:per-language} per-language escalation, and
Appendix~\ref{app:err} the layered controls that bound under-escalation. Together with the steady production
handoff rate (Section~\ref{sec:ab}), these breakdowns let a reader check that
the escalation reduction does not come from under-escalation.
Cost and infrastructure
figures are illustrative estimates, not audited financial statements or actuals.
\bibliography{main}
\appendix
\section{Context-Contract Specification}
\label{app:ctx}
The forward contract restricts what the generator can see and reference; the return
contract restricts what it can send back to the backend. The tables use the full
page width because field descriptions are part of the interface specification.
The contract enforces four invariants. First, reference closure: every article ID,
structured action, and user-state key in the return object must be a member of the
forward object. Second, role separation: the generator may explain an action but may
not choose a different action path. Third, replayability: the forward object plus
tool observations is sufficient to reconstruct why the user-facing response was
allowed. Fourth, localization safety: locale is carried as data, while translation
and rendering stay in backend services rather than being delegated to free-form
model text. These invariants are checked by backend code, not by trusting the model
to follow natural-language instructions alone.
\begin{table*}[t]
\centering\small
\setlength{\tabcolsep}{4pt}
\begin{tabularx}{\textwidth}{@{}
>{\raggedright\arraybackslash}p{4.7cm}
>{\raggedright\arraybackslash}p{1.55cm}
>{\raggedright\arraybackslash}X
@{}}
\toprule
Field & Type & Description \\
\midrule
\texttt{task\_state} & enum & Orchestrator objective: \texttt{answer\_informational}, \texttt{present\_action\_card}, \texttt{ask\_clarifying}, \texttt{escalate}, or \texttt{fallback}. \\
\texttt{available\_tools} & list & Structured actions and tools eligible for this turn; the generator may reference only these members. \\
\texttt{rag\_context} & list & Retrieved KB or policy passages with \texttt{source\_id}, title, and snippet. \\
\begin{tabular}[t]{@{}l@{}}\texttt{contextual\_}\\\texttt{variables}\end{tabular} & map & User-state variables such as reservation, listing, payment, and account state, keyed for citation. \\
\texttt{prior\_decisions} & list & Earlier tool calls and observations in the bounded ReAct loop. \\
\texttt{policy\_constraints} & list & Policy and risk constraints the reply must respect. \\
\texttt{locale} & string & User language and region for tone and translation. \\
\bottomrule
\end{tabularx}
\caption{Forward context contract: fields the orchestrator passes to the generator.}
\label{tab:fwd}
\end{table*}
\begin{table*}[t]
\centering\small
\setlength{\tabcolsep}{4pt}
\begin{tabularx}{\textwidth}{@{}
>{\raggedright\arraybackslash}p{4.7cm}
>{\raggedright\arraybackslash}p{1.55cm}
>{\raggedright\arraybackslash}X
@{}}
\toprule
Field & Type & Description \\
\midrule
\texttt{response\_text} & string & User-facing reply, pre-translation. \\
\texttt{resolution\_path} & enum & Canonical support path selected by the orchestrator. \\
\texttt{cited\_article\_ids} & list & KB/policy IDs used by the reply; the backend verifies membership in \texttt{rag\_context}. \\
\texttt{cited\_cv\_keys} & list & User-state-variable keys referenced; verified against \texttt{contextual\_variables}. \\
\texttt{action\_card\_ref} & optional & Echo of the orchestrator-selected structured action to render; rejected if absent from \texttt{available\_tools} or inconsistent with the orchestrator decision. \\
\texttt{contextual\_information} & map & Renderable fields such as dates, amounts, and names. \\
\texttt{handoff\_status} & enum & Backend status for normal generation, clarification, fallback, or escalation explanation. \\
\bottomrule
\end{tabularx}
\caption{Return context contract: fields the generator returns to the backend.}
\label{tab:ret}
\end{table*}
\section{Additional Error and Component Analysis}
\label{app:err}
This appendix expands process evidence from Section~\ref{sec:comparison}; example
conversations are synthetic. Table~\ref{tab:evidence_scope} summarizes which
instrument supports which claim; Table~\ref{tab:attribution} (main text)
attributes each main gain to its cause.
\begin{table}[htbp]
\centering
\scriptsize
\setlength{\tabcolsep}{3pt}
\renewcommand{\arraystretch}{0.94}
\begin{tabular}{@{}p{0.25\linewidth}p{0.22\linewidth}p{0.43\linewidth}@{}}
\toprule
Evidence & Unit / source & Use in this paper \\
\midrule
Shadow replay & Turn; fixed replay traffic & Compare systems without exposure and localize component failures. \\
Tool-decision checks & Tool call; labeled slices & Measure selector, structured-action, and escalation decisions. \\
Human/risk review & Conversation; dual labels & Gate broadening and identify risk modes. \\
Online A/B & User/session; low-ramp RCT & Check escalation, contact, and self-solve trends; report confidence where powered. \\
Serving logs & Request & Verify latency budget and GPU footprint. \\
\bottomrule
\end{tabular}
\caption{Release evidence for the architecture comparison. Instrument construction,
judge certification, and offline-to-online calibration are not contributions.}
\label{tab:evidence_scope}
\end{table}
\begin{table*}[t]
\centering\small
\setlength{\tabcolsep}{5pt}
\begin{tabular}{@{}llrrrrrll@{}}
\toprule
Tier & Escalation & $n_C$ & $n_T$ & Control & Treatment & $\Delta$ (pp) & $p$ & 95\% CI (pp) \\
\midrule
English & hard & 7,824 & 7,830 & 5.24\% & 2.80\% & $-2.44$ & $7.1 \times 10^{-15}$ & $[-3.06, -1.83]$ \\
English & soft & 7,824 & 7,830 & 9.43\% & 2.03\% & $-7.40$ & $2.6 \times 10^{-88}$ & $[-8.12, -6.68]$ \\
Tier 1 (ES, FR) & hard & 1,085 & 1,203 & 5.07\% & 3.08\% & $-1.99$ & $0.015$ & $[-3.62, -0.36]$ \\
Tier 1 (ES, FR) & soft & 1,085 & 1,203 & 9.12\% & 2.00\% & $-7.13$ & $4.4 \times 10^{-14}$ & $[-9.02, -5.24]$ \\
\quad ES & hard & 931 & 1,024 & 4.83\% & 3.22\% & $-1.61$ & $0.069$ & $[-3.36, +0.14]$ \\
\quad ES & soft & 931 & 1,024 & 9.02\% & 1.86\% & $-7.16$ & $1.4 \times 10^{-12}$ & $[-9.18, -5.14]$ \\
\quad FR & hard & 154 & 179 & 6.49\% & 2.23\% & $-4.26$ & $0.054$ & $[-8.71, +0.19]$ \\
\quad FR & soft & 154 & 179 & 9.74\% & 2.79\% & $-6.95$ & $0.008$ & $[-12.22, -1.68]$ \\
Overall & hard & 8,926 & 9,073 & 5.23\% & 2.82\% & $-2.41$ & $1.8 \times 10^{-16}$ & $[-2.98, -1.84]$ \\
Overall & soft & 8,926 & 9,073 & 9.41\% & 2.03\% & $-7.38$ & $2.1 \times 10^{-101}$ & $[-8.05, -6.71]$ \\
\bottomrule
\end{tabular}
\caption{Per-language-tier hard and soft escalation: turn-level response
classifications on evaluated turns over 2026-03-05 to 2026-03-12, the widest
window in which both arms served non-shadow traffic. $\Delta$ is treatment
minus control in points. The Overall rows additionally include a small number
of turns not attributed to a listed language tier (17 control, 40 treatment),
so they slightly exceed the sum of the English and Tier 1 rows. English and
Tier 1 confidence intervals overlap on
both metrics, indicating no language-tier interaction. Within Tier 1, both
single-language soft-escalation reductions are significant, while the
single-language hard-escalation deltas are directional at these sample sizes
(their CIs cross zero); ES turn counts are derived as Tier 1 minus FR, since
Portuguese carried no assistant traffic in this window, and the ES/FR
confidence intervals are computed with the same unpooled two-proportion method
as the tier rows. Tier 2 (DE, IT, RU, JA, KO, TH, ZH) was
routed to the legacy bot during the experiment and carried no assistant
traffic in either arm; the 11 languages in the abstract describe the deployed
platform, not this experiment's eligible population. Session-level results
over the full experiment window (2026-02-24 to 2026-04-14; $n \approx 139$K
engaged sessions per arm, a different denominator) show the same consistency:
self-solve $+1.77$ points for English and $+1.80$ for Tier 1, combined
escalation $-1.68$ and $-1.77$ (all $p < 10^{-8}$), and end-to-end P90 latency
improves from 9.43s to 7.34s for English and 10.08s to 8.25s for Tier 1,
though Tier 1 remains slower than English in both arms. Tier-level differences
are not driven by language routing: the language-identification guardrail
exceeds 99\% weighted accuracy on human-labeled production data.}
\label{tab:per-language}
\end{table*}
\paragraph{Per-category under-escalation controls.}
Under-escalation is bounded by layered controls rather than by orchestrator
discretion alone. First, safety-sensitive content is detected by dedicated
per-category guardrail models (safety, jailbreak, inappropriate content) and
routed deterministically to humans, independent of the orchestrator decision.
The safety detector is evaluated offline against human labels and operated at
a deliberately recall-first point that trades false positives for coverage of
policy-relevant safety content, so detector misses --- the per-category
under-escalation channel --- are the minority case by design. Jailbreak and
inappropriate-content guardrails run on the same independent routing path and
are threshold-calibrated on production traffic against judge-labeled samples.
The detectors' operating points are security-sensitive and not disclosed.
Second, the user can request human escalation at any turn, and user-initiated
handoff volume held roughly steady across the migration
(Section~\ref{sec:ab}), so the response-level escalation reduction did not
suppress user-initiated handoffs. Third, a post-interaction outreach program
reviews flagged conversations and follows up after the fact, so a missed live
escalation is not a terminal failure. Production prevalence of these
categories is low (well under 0.1\% of turns for all but general safety
topics), so per-category under-escalation rates carry wide confidence
intervals and are monitored operationally rather than estimated from the
shadow window.
\paragraph{Structured-action selection.}
Table~\ref{tab:ac} reports precision, recall, F1, and false-positive proportion
across routing-prompt iterations. DR reaches F1 parity while making the operating
point explicit.
\begin{table}[htbp]
\centering\small
\setlength{\tabcolsep}{5pt}
\begin{tabular}{lcccc}
\toprule
System / prompt & P & R & F1 & FPP \\
\midrule
Monolithic (prod) & 80.7 & 75.2 & 77.9 & 6.6 \\
DR (v1) & 82.5 & 70.5 & 76.0 & \textbf{5.5} \\
DR (v2) & 77.9 & 77.4 & 77.7 & 8.0 \\
\bottomrule
\end{tabular}
\caption{Structured-action precision, recall, F1, and false-positive proportion
(\%). DR reaches F1 parity; the routing prompt sets the operating point.}
\label{tab:ac}
\end{table}
\paragraph{Human-review error taxonomy.}
\begin{table}[htbp]
\centering\small
\setlength{\tabcolsep}{4pt}
\begin{tabular}{@{}lcl@{}}
\toprule
Error mode & Share & Denominator \\
\midrule
Intent classification & 34.8\% & all errors \\
Missed escalation trigger & 30.4\% & all errors \\
Hallucination (gap-filling) & 55.0\% & groundedness errors \\
\bottomrule
\end{tabular}
\caption{Top error modes from human review. Shares use different denominators and
are not a mutually exclusive partition.}
\label{tab:errtax}
\end{table}
\paragraph{Soft-escalation buckets.}
Of the Monolithic soft escalations that DR resolves without escalation
(approximately 3\% of total traffic), Table~\ref{tab:buckets} shows the scenario
mix used for policy review.
\begin{table}[htbp]
\centering\small
\setlength{\tabcolsep}{6pt}
\begin{tabular}{@{}lc@{}}
\toprule
Scenario & Share \\
\midrule
Technical error & 29.3\% \\
Booking modification / cancellation & 17.0\% \\
Check-in / access & 12.9\% \\
Payment / billing & 12.2\% \\
Unresponsive host & 8.8\% \\
Other (long tail) & 19.8\% \\
\bottomrule
\end{tabular}
\caption{Scenario mix of soft escalations resolved by DR without escalation.
``Other'' is the residual after the named buckets.}
\label{tab:buckets}
\end{table}
\paragraph{Generator user-state-variable attribution.}
The elevated generator-side USV attribution error in Table~\ref{tab:turn} comes
from strict attribution and genuine gap filling. In the first case, a reply may
correctly summarize multiple variables but cite only one key; in the second, the
generator states a value absent from the contract. We are moving from
per-sentence to per-key attribution to separate these cases. Gap filling is also structurally bounded:
post-generation checks reject any user-state key outside the contract, so the
residual exposure is wrong values for legitimate keys, which per-key attribution
measures directly.
\paragraph{Illustrative failures.}
Ambiguous requests such as ``I need to change my trip'' can trigger the wrong
policy instead of a clarifying question. Safety-sensitive check-in and access
issues can also require escalation even when a help article exists. Both cases are
handled as routing-policy updates rather than generator-only fixes.
\section{Component Design and Optimization Details}
\label{app:components}
This appendix details design choices behind DR. It reports operational component
checks, not a new measurement methodology; scoring-instrument construction and
production-outcome validation remain out of scope. The intent is to document the
interfaces and choices a reader would need to reproduce the decomposition: what the
planner can choose, what context it receives, how routing prompts were isolated, and
why the smaller generator was acceptable. Throughout, \textbf{L} denotes
Qwen3-235B-A22B for orchestration and \textbf{S} denotes Qwen3-Next-80B-A3B for
generation.
\subsection{Architectural decomposition}
\label{app:comp-decomp}
Monolithic fixed much of the turn flow inside a single blended model path over
preassembled context. DR instead lets the orchestrator choose response
generation, escalation, reservation and listing selectors, structured-action
retrieval, help-content retrieval, and user-state-variable retrieval. The
generator receives the selected path through the contract and cannot choose a
different action. This exposes separately tunable surfaces: tool schemas,
selector context, routing prompts, generator prompts, fallback behavior, and
model placement. The registry is versioned with the prompts: when a tool schema
changes, both the orchestrator prompt and backend validator are advanced
together, then shadowed before user exposure. Table~\ref{tab:tools} lists the
full registry referenced in Section~\ref{sec:dr}.
\begin{table*}[t]
\centering
\begingroup
\scriptsize
\setlength{\tabcolsep}{2.5pt}
\renewcommand{\arraystretch}{0.86}
\begin{tabular}{@{}p{0.18\textwidth}p{0.12\textwidth}p{0.25\textwidth}p{0.39\textwidth}@{}}
\toprule
Tool class & Visibility & Output to orchestrator & Architectural purpose \\
\midrule
USV retrieval & Internal & User-state variables under stable keys & Provides reservation, listing, payment, and account state without exposing the full user record. \\
KB retrieval & Internal & Ranked policy/help snippets with source IDs & Grounds informational answers and constrains citation targets. \\
Structured-action retrieval & Internal & Eligible action-card subset & Limits actions to cards renderable for this user and turn. \\
Reservation/listing selectors & Internal or user-facing & Selected entity ID or null decision & Prevents the generator from guessing the referenced entity. \\
Escalation & User-facing & Hard-route decision and reason code & Makes human routing an explicit action, not generated phrasing. \\
Response generation & User-facing & Final response from generator & Runs only after task state and context are fixed. \\
Fallback generation & User-facing & Conservative response over available context & Handles ReAct iteration limit or invalid orchestration state. \\
\bottomrule
\end{tabular}
\endgroup
\caption{Orchestrator tool registry. Each tool has a typed schema, eligibility constraints, and logging fields.}
\label{tab:tools}
\end{table*}
\subsection{Selector context implementation}
\label{app:selector-context}
The selector input contains the set of candidate reservation or listing IDs,
whether an ID is currently active in the conversation, whether it came from the
user message, profile state, itinerary state, or a previous tool observation, and
a short natural-language mention span when one exists. The selector can return a
concrete ID, a \texttt{null} decision when no entity is needed, or a clarification
request when multiple entities remain plausible. This schema moved entity choice
out of final prose: the generator no longer has to infer a reservation from
wording, and the backend can reject any rendered card whose ID was not selected by
the orchestrator.
\subsection{Planner model selection}
\label{app:planner-model}
The orchestrator must produce short but high-impact routing decisions. We compared
candidate planners on action-card selection against the same human-labeled set
(Table~\ref{tab:planner-model}). Among untuned candidate planners, GPT-4o had the
strongest initial F1, but \textbf{L} was already deployed, self-hosted, and
adaptable in house; we selected \textbf{L} and closed the gap through
routing-prompt iteration.
\begin{table*}[t]
\centering
\small
\begin{tabular}{lcccc}
\toprule
Planner model (initial prompt) & Precision & Recall & F1 & FPP \\
\midrule
Production prompt on \textbf{L} (reference) & 0.807 & 0.752 & 0.779 & 0.066 \\
GPT-4o & 0.782 & 0.760 & 0.771 & 0.077 \\
In-house Qwen3-235B-A22B (\textbf{L}) & 0.701 & 0.714 & 0.707 & 0.111 \\
Vanilla Qwen3-235B (base) & 0.728 & 0.497 & 0.591 & 0.068 \\
Vanilla Qwen3-235B (instruct) & 0.693 & 0.705 & 0.699 & 0.114 \\
\bottomrule
\end{tabular}
\caption{Planner model comparison for action-card selection against the same
human-labeled set. FPP denotes false-positive proportion.}
\label{tab:planner-model}
\end{table*}
\subsection{Action-card routing prompt}
\label{app:ac-routing}
After selecting \textbf{L}, we tuned the action-card routing prompt. The first DR
prompt raised recall but over-fired; adding selector context raised recall further;
later instructions reduced over-firing while preserving most recall (Table~\ref{tab:ac-routing}).
\begin{table*}[t]
\centering
\footnotesize
\setlength{\tabcolsep}{4pt}
\begin{tabular}{lccccc}
\toprule
Planner prompt & Response-tool rate & Precision & Recall & F1 & FPP \\
\midrule
Production prompt (reference) & n/a & 0.807 & 0.752 & 0.779 & 0.066 \\
DR V1.0 & 93.8\% & 0.712 & 0.831 & 0.767 & 0.122 \\
DR V1.1 + picker context & 97.6\% & 0.695 & 0.863 & 0.770 & 0.138 \\
DR V1.2.4 + picker context & 97.0\% & 0.752 & 0.816 & 0.783 & 0.098 \\
\bottomrule
\end{tabular}
\caption{Action-card routing against the human golden label, planner on
\textbf{L}. Later prompt iterations explored the F1 and false-positive trade-off;
Appendix~\ref{app:err} identifies the gate-compliant deployed point.}
\label{tab:ac-routing}
\end{table*}
\subsection{Help-content routing}
\label{app:help-content-routing}
The help-content path uses a stage-aware primary-source rule: ``why'' or ``can I''
queries prefer policy or overview articles, while ``how'' or ``what do I do''
queries prefer procedural articles. Encoding this at routing time keeps source
choice separate from wording and improves labeled help-content selection
(Table~\ref{tab:help-content}). The generator then explains the selected source
rather than searching again, which prevents late-stage source drift.
\begin{table*}[t]
\centering
\small
\begin{tabular}{lccc}
\toprule
System & Precision & Recall & F1 \\
\midrule
Monolithic reference & 32.32\% & 75.44\% & 34.15\% \\
DR + routing-prompt tuning & 37.25\% & 85.32\% & 41.39\% \\
\bottomrule
\end{tabular}
\caption{Help-content selection against ground-truth labels on approximately 800
samples. F1 is macro-averaged per-class F1, so it does not equal the harmonic
mean of the aggregate precision and recall columns.}
\label{tab:help-content}
\end{table*}
\subsection{Generator model selection}
\label{app:generator-model-selection}
The generator must leave latency headroom for orchestration, retrieval, guardrails,
and translation. We compared the larger \textbf{L}, the smaller \textbf{S}, and
frontier APIs on a production-shaped synthetic set (Table~\ref{tab:generator-models}).
Claude Opus 4 was infeasible on latency in this setup. We chose \textbf{S} because
it met release thresholds at lower P90 and about half the serving footprint of
\textbf{L}. Selecting a smaller model for the narrower generation role is
consistent with model compression and on-policy distillation results
\citep{hinton2015distill,agarwal2024gkd}; here we use an existing open-weight
model aligned to policy (Appendix~\ref{app:alignment}) rather than distilling our
own.
\begin{table*}[t]
\centering
\footnotesize
\setlength{\tabcolsep}{4pt}
\begin{tabular}{llccc}
\toprule
Generation model & Serving footprint & P90 latency & Correctness & Groundedness \\
\midrule
\textbf{L}: Qwen3-235B-A22B & 2$\times$ & 6.29 s & 93.3\% & 86.6\% \\
\textbf{S}: Qwen3-Next-80B-A3B (deployed) & 1$\times$ (reference) & 5.45 s & 92.1\% & 83.1\% \\
GPT-4.1 (API) & n/a & 4.76 s & 94.4\% & 89.4\% \\
Claude Opus 4 (API) & n/a & 17.81 s & 94.8\% & 87.2\% \\
\bottomrule
\end{tabular}
\caption{Generator candidates on a production-shaped synthetic set: serving
footprint (relative to the deployed configuration), model-only P90 latency, and
release-check scores. Latency is measured under model-selection settings and is
not directly comparable to the production generation-stage latency in
Table~\ref{tab:e2e-latency}.}
\label{tab:generator-models}
\end{table*}
\subsection{Generator prompt and residual headroom}
\label{app:generator-prompt-headroom}
With \textbf{S} selected, we compared a reduced-context per-solution prompt with a
merged full-context prompt. The reduced prompt is attractive because it passes only
the action-card, help-content, or contextual-variable slice selected by the
orchestrator. The merged prompt is operationally simpler: it always passes the same
schema and lets the generator decide how much of the allowed context to mention. The
core quality scores were close, so we chose the merged prompt because one schema
removes per-path branching, shortens replies, and slightly improves style, tone, and
repetition behavior (Table~\ref{tab:generator-prompt}).
\begin{table*}[t]
\centering
\footnotesize
\setlength{\tabcolsep}{4pt}
\begin{tabular}{lcccccc}
\toprule
Generator prompt & Resolution & Groundedness & Correctness & Repeat-instr. $\downarrow$ & Words & Style/tone \\
\midrule
Reduced-context (per-solution) & 0.84 & 0.84 & 0.91 & 0.24 & 109 & 0.65 \\
Full-context (merged, deployed) & 0.81 & 0.83 & 0.92 & 0.21 & 61 & 0.71 \\
\bottomrule
\end{tabular}
\caption{Generator prompting regimes on \textbf{S} over a shared shadow-traffic set
of approximately 455 samples. Scores are in $[0,1]$ except word count; lower is
better for repeated instruction.}
\label{tab:generator-prompt}
\end{table*}
Table~\ref{tab:generator-headroom} shows the residual gap between \textbf{S} and
larger \textbf{L}. We treat it as future generator headroom, not evidence against
the split, because the architecture gains come from better upstream decisions.
\begin{table*}[t]
\centering
\footnotesize
\setlength{\tabcolsep}{4pt}
\begin{tabular}{lccccc}
\toprule
Generation model & Resolution & Groundedness & Correctness & Words & Style/tone \\
\midrule
\textbf{S}: deployed full-context generator & 0.81 & 0.83 & 0.92 & 61 & 0.71 \\
\textbf{L}: larger in-house model & 0.86 & 0.85 & 0.94 & 131 & 0.68 \\
$\Delta$ (\textbf{S} $-$ \textbf{L}) & $-$0.05 & $-$0.02 & $-$0.02 & $-$70 & $+$0.03 \\
\bottomrule
\end{tabular}
\caption{Deployed generator \textbf{S} versus larger in-house \textbf{L} on a shared
shadow-traffic set. $\Delta$ is \textbf{S} minus \textbf{L}.}
\label{tab:generator-headroom}
\end{table*}
\section{Serving Model Detail}
\label{app:model}
\paragraph{Speculative decoding and co-location.}
EAGLE-3 speculative decoding is being validated as a serving-only optimization
for a further-reduced footprint target (Table~\ref{tab:latency}). The draft
model fuses low, middle, and high
verifier-layer hidden states to predict the next three tokens; under standard
speculative verification, accepted outputs preserve the verifier distribution.
We train the draft on production-shaped traces, warm-start from a public
Qwen3-235B speculator checkpoint, and validate on shadow traffic so the
optimization changes serving rather than behavior. We co-locate orchestrator and
generator capacity so the two share one availability envelope; future generator
swaps leave the orchestrator interface unchanged because the contract is stable.
Table~\ref{tab:latency} reports orchestrator-only serving latency across operating
points, expanding the headline figures in Section~\ref{sec:serving}. Footprints
are stated relative to the pre-optimization baseline.
\begin{table}[htbp]
\centering\small
\setlength{\tabcolsep}{4pt}
\begin{tabular}{lcccc}
\toprule
Config & P90 & P95 & P99 & GPU $\Delta$ \\
\midrule
Baseline & 3.87 & 4.53 & 5.95 & n/a \\
Opt.~1 & 2.24 & 2.61 & 3.43 & $-33\%$ \\
Opt.~2 & 2.60 & 3.02 & 3.87 & $-56\%$ \\
Opt.~3 & 2.66 & 3.16 & 4.07 & $-67\%$ \\
\bottomrule
\end{tabular}
\caption{Orchestrator serving latency in seconds and GPU-footprint reduction
relative to the pre-optimization baseline. Opt.~1 is the current production
operating point; Opt.~2 is the rollout target for speculative decoding; Opt.~3
is a candidate operating point that trades slightly higher latency for a
further-reduced footprint.}
\label{tab:latency}
\end{table}
Table~\ref{tab:models} summarizes orchestrator candidates from a serving and
controllability perspective: active parameters, latency under our stack, and whether
the model allows in-house adaptation. We selected Qwen3-235B-A22B not because it was
the highest-quality model on every slice, but because it gave the best joint
operating point: low served latency in our stack, self-hosted control, and a path to
future in-house adaptation without changing the tool interface.
\begin{table}[htbp]
\centering\scriptsize
\setlength{\tabcolsep}{3.5pt}
\begin{tabular}{lcccl}
\toprule
Model & Act. prm & P50 & P90 & Control \\
 &  & (s) & (s) &  \\
\midrule
Claude Opus 4.5 & n/a & 3.19 & 4.41 & API only \\
GPT-4.1 & n/a & 1.86 & 2.42 & API only \\
Gemini 3 Pro & n/a & 12.12 & 26.11 & API only \\
Gemini 3 Flash & n/a & 2.25 & 3.16 & API only \\
Claude Haiku 4.5 & n/a & 3.15 & 4.41 & API only \\
Kimi 2.5 & 32B & 5.46 & 8.73 & open weights \\
Qwen3-235B-A22B (ours) & 22B & 1.56 & 1.96 & self-hosted \\
\bottomrule
\end{tabular}
\caption{Orchestrator model options from a serving and controllability perspective.
Latency is measured on the model-selection benchmark under internal settings and
is not directly comparable to the production orchestrator latency in
Table~\ref{tab:latency}. Proprietary models are API-served, while Qwen3 is
self-hosted.}
\label{tab:models}
\end{table}
\section{Alignment Data and Reward Abstraction}
\label{app:alignment}
This appendix gives a minimal description of the alignment stage used after
model selection. We intentionally keep implementation details abstract because
this paper focuses on the deployed orchestrator and generator interface, not on a
full evaluation or reward-calibration study.
\subsection{Alignment data}
\label{app:align-data}
We construct alignment examples from de-identified production and simulated
support cases after removing opt-out traffic. Cases are sampled to cover both
successful and unsuccessful interactions. For each input, the current policy
produces multiple candidate responses under the same conversation state,
retrieved context, and available actions. This creates within-input comparisons
that isolate response-policy behavior while holding the orchestrator inputs
fixed.
The resulting candidates are scored by task rewards and converted into the
training format required by the chosen post-training method, following the
RLHF lineage \citep{ouyang2022instructgpt,schulman2017ppo}. Pairwise methods use
higher-scoring and lower-scoring responses as preference pairs
\citep{rafailov2023dpo,hong2024orpo}, while relative-reward methods use the
candidate set to compute group-level advantages
\citep{shao2024grpo,zheng2025gspo}.
\subsection{Reward abstraction}
\label{app:align-reward}
We use two reward families. A quality-centered reward scores whether the response
is correct, grounded, clear, and compliant with the prompt. A reference-aware
reward additionally checks whether the response covers a likely solution and uses
supporting knowledge-base or action-card evidence consistently. The second family
is intended to discourage safe but under-resolving answers, such as generic
policy summaries or clarification-only responses when enough evidence exists to
provide a conditional next step.
To reduce leakage between training and evaluation, reward scoring is separated
from final offline evaluation. Before serving, the selected checkpoint is frozen
and evaluated through the same deployment checks as other candidate updates. The
serving interface is unchanged: the orchestrator still supplies conversation
state, retrieved context, and available actions, while only the generator policy
is replaced.
\section{Serving-Cost Estimate}
\label{app:cost}
For illustration only, we estimate annual model-serving cost at our volume and
decoding settings at roughly \$6M for the proprietary-API baseline versus
roughly \$200K self-hosted. Both are order-of-magnitude internal estimates,
not audited financials or actuals. We state the basis so readers can rescale
to their own setting.
\paragraph{Proprietary-API baseline.}
Estimated from annual generated-plus-prompt token volume across orchestration and
generation at production traffic, and the blended list price of the proprietary
API considered. For rescaling: the orchestrator processes roughly 8,000 prompt and 200 completion
tokens per model call (up to three calls per turn), and annual turn
volume is on the order of $10^{7}$.
\paragraph{Self-hosted.}
Estimated as amortized hardware plus power for the production footprint
(the orchestrator at its current production operating point plus co-located
generator capacity), at our measured utilization over a standard amortization
period. It \emph{excludes} engineering and on-call labor, model-development
cost, and shared-platform overhead; including loaded labor would raise the
self-hosted figure but not by enough to close an order-of-magnitude gap.
\paragraph{Scope of the estimate.}
The comparison is therefore self-hosted hardware cost against API list price. The
takeaway we stand behind is a greater-than-tenfold reduction in \emph{serving}
cost at our scale, not a precise ratio. The figures move with traffic, hardware
generation, utilization, and API pricing.
\end{document}